\documentclass{article} 

\usepackage{times}
\usepackage{soul}
\usepackage{url}
\usepackage{graphicx}
\usepackage{amsmath}
\usepackage{amsthm}
\usepackage{booktabs}
\usepackage{longtable}
\usepackage{xcolor}
\usepackage{colortbl} 
\usepackage{algorithm}
\usepackage{colortbl}
\usepackage{booktabs} 
\usepackage{multirow} 
\usepackage{tabularx} 
\usepackage{geometry} 
\usepackage{pdflscape}
\usepackage{rotating} %
\usepackage{authblk}  
\usepackage{hyperref} 
\graphicspath{{Fig/}}

\hypersetup{
    colorlinks=true,
    linkcolor=blue,       
    citecolor=blue,     
    filecolor=magenta,   
    urlcolor=cyan        
}

\definecolor{babypink}{rgb}{0.96, 0.76, 0.76}
\definecolor{babyblue}{rgb}{0.54, 0.81, 0.94}
\definecolor{bananamania}{rgb}{0.98, 0.91, 0.71}

\theoremstyle{thmstyleone}%
\theoremstyle{thmstyletwo}%
\theoremstyle{thmstylethree}%

\renewenvironment{abstract}{
  \begin{center}
    \normalfont\normalsize\bfseries Abstract\vspace{-.5em}\vspace{0pt}
  \end{center}
  \quote\normalsize
}{\endquote}

\title{Recent Developments in GNNs for Drug Discovery}

\author[1]{Zhengyu Fang}
\author[1]{Xiaoge Zhang}
\author[1]{Anyin Zhao}
\author[1,2,3,4]{Xiao Li}
\author[1]{Huiyuan Chen}
\author[1]{Jing Li\thanks{Corresponding author. \href{mailto:jingli@case.edu}{jingli@case.edu}}}

\affil[1]{Department of Computer and Data Sciences, Case Western Reserve University}
\affil[2]{Department of Biochemistry, Case Western Reserve University}
\affil[3]{Center for RNA Science and Therapeutics, Case Western Reserve University}
\affil[4]{Department of Biomedical Engineering, Case Western Reserve University}

\date{}  

\begin{document}

\maketitle

\begin{abstract}
In this paper, we review recent developments and the role of Graph Neural Networks (GNNs) in computational drug discovery, including molecule generation, molecular property prediction, and drug-drug interaction prediction. By summarizing the most recent developments in this area, we underscore the capabilities of GNNs to comprehend intricate molecular patterns, while exploring both their current and prospective applications. We initiate our discussion by examining various molecular representations, followed by detailed discussions and categorization of existing GNN models based on their input types and downstream application tasks. We also collect a list of commonly used benchmark datasets for a variety of applications. We conclude the paper with brief discussions and summarize common trends in this important research area.
\end{abstract}

\vspace{1em}
\noindent \textbf{Keywords:} Drug Discovery, Graph Neural Networks, Machine Learning

\section{Introduction}
It is well known that traditional drug discovery is costly, time-consuming, and with high failure rates~\cite{mullard2014new}. To streamline the process of drug discovery and mitigate resource-intensive laboratory work, significant research has been dedicated to the development of computational methods. Existing literature provides some comprehensive reviews on deep learning approaches in drug discovery~\cite{gawehn2016deep,chen2019modeling,chen2018drugcom, wang2014drug}. In this review, we focus on the development and applications of Graph Neural Networks (GNNs) on three related areas of computational drug development, namely, \textit{Molecule Generation}, \textit{Molecular Property Prediction}, and \textit{Drug-Drug Interaction Prediction}, which not only receive increasing attention but also show promising results. We will summarize some most recent developments in these research areas and focus on computational advances published since 2021. 


\subsection{Molecule Generation}
An earlier step in developing a drug for a certain disease usually involves the search of  enormous amount of molecules to create a small subset of candidates for further laboratory testing. To reduce experimental costs and to produce drug candidates more efficiently, many computational approaches have been developed to generate novel molecules~\cite{xu2021endtoend,swanson2023von,maziarz2022learning,geng2023novo,anonymous2024drug}. These approaches usually take some existing molecules as inputs and generate valid but different molecules with high variety. Validity basically means that the generated molecules are possible to be synthesized. At the same time, high variety will ensure that the generated molecules span over a reasonably large molecular space in hope that some of them will have desired properties. Computational approaches are usually evaluated based on these two metrics.  

Generally speaking, existing computational approaches can be characterized as two different types based on whether they can incorporate certain constrains. For some approaches~\cite{xu2021endtoend,swanson2023von}, any molecules can be generated, \textit{i.e.},  the generation of molecules has no constraints associated with them. For the second type of approaches~\cite{maziarz2022learning,geng2023novo,anonymous2024drug}, they only generate molecules satisfying certain constrains. For example, all the generated molecules must include some fixed substructures such as ring structures. 
In other cases, the constraints can be task specifics, for example, many approaches~\cite{luo20213d,liu2022generating,peng2022pocket2mol,zhang2023molecule,adams2022equivariant} have been developed to generate molecules that bind to certain protein binding sites. 



Early computational approaches for molecule generation were mostly applications of generative deep learning models, such as variational autoencoder~\cite{Lim2018}. More recently, there was a shift from generations of whole molecules all at once, to iterative generations of atoms and bonds~\cite{xu2021endtoend, maziarz2022learning}. Many of such approaches utilized GNN modules as backbones to generate molecules with specific substructures or with high binding affinity to specific proteins~\cite{maziarz2022learning}.

\subsection{Molecular Property Prediction}
Once drug candidates are selected, researchers need to study drugable properties of candidate molecules, such as toxicity and possible adverse effects, water solubility, and binding affinity to target proteins. Experimental studies are slow and costly. With quick accumulation of high-quality training data, a great number of computational methods for molecular property prediction have been developed~\cite{chen2020learning,sun2021mocl,ju2023few,anonymous2024pacia}, the goals of which are to reduce the need for extensive experimental validation and to accelerate the drug discovery process. 

Molecular properties can be classified into different categories based on their levels, from molecular level (e.g., quantum mechanics) to physiological level~\cite{wu2018moleculenet}. Computationally, the tasks of molecular property prediction can be categorized into two broad classes based on their inputs: predictions based on single molecules (i.e., drug candidates)~\cite{chen2020learning,yang2021deep,zhu2022mathcal}, and predictions of relationships based on two or more molecules (e.g., drug-target or protein-ligand)~\cite{jones2021improved,jin2023unsupervised,zhang2022e3bind}. The predictions based on single molecules can be either classification-based or regression-based. The goal of classification tasks is to predict the presence/absence of certain properties. The regression tasks focus on the prediction of the quantitative value of a particular property of a molecule. Similarly, relationships among molecules~\cite{yang2022mgraphdta,lee2023conditional} can be represented by a binary value or a numerical value. Examples of the former include whether a protein and a target interact. The affinity score between two molecules is an example of the latter.

Traditionally, drug property prediction heavily relied on classical machine learning techniques such as random forests or support vector machines~\cite{gawehn2016deep}, necessitating deep domain knowledge for feature engineering. With the emergence of deep learning, there was a significant shift towards neural network based models~\cite{gawehn2016deep,chen2020idrug}. This transition marked a move from one-dimensional string representations of drugs to richer two-dimensional graph and three-dimensional conformation models, broadening the scope and accuracy of drug property prediction. In particular, GNNs, which take in a molecular graph representation and directly predict molecular properties, have attracted substantial attention in cheminformatics and bioinformatics (\textit{e.g.}, ~\cite{stark20223d}).

\subsection{Drug-Drug Interaction Prediction}
In the treatment of complex diseases such as cancer and neurological disorders, combinational drug therapies have shown promise in improving treatment efficacy by targeting multiple biological pathways simultaneously~\cite{chen2018drugcom,menden_community_2019,preuer2020leveraging}. However, as the number of possible drug combinations grows with the increasing number of available drugs, the potential for undetected and unexpected drug-drug interaction (DDIs) also rises~\cite{vilar2014similarity}. Such interactions can lead to reduction of therapeutic effectiveness, adverse side effects~\cite{jiang2022adverse}, and sometimes increased hospitalization~\cite{dechanont2014hospital}, posing significant challenges to the screening and optimization of combinational drug therapies. Therefore, it is crucial to develop methods that can precisely predict the drug-drug synergies and adverse interactions to ensure safer and more effective treatments.

Strictly speaking, drug combination therapies, which rely on synergistic or additive interactions of drugs to increase efficacy, reduce toxicity, and/or prevent drug resistance,  can be treated as one type of drug-drug interactions. In this review, we use the broad definition of drug-drug interactions, which include both synergistic/additive as well as antagonistic/adverse interactions. To predict drug-drug interactions computationally, the inputs usually include representations of drugs and types of interactions. In the simplest case, methods may only focus on the prediction of the existence of a particular type of interaction~\cite{wang2022predicting}. The problem is simply formulated as a binary classification problem. In other cases (\textit{e.g.}, data from cell-based assays), the relationships can be quantified based on one or multiple measures of individual drugs and/or drug pairs (e.g., half maximal inhibitory concentration or IC50), which can be viewed as regression tasks. Yet in some other cases (data from patient-based studies), different criteria or biomarkers (\textit{e.g.}, blood pressure, blood sugar) are considered and only a few categorical values (\textit{e.g.}, increasing, no change, decreasing) are considered for each criterion/biomarker. The problem can be formulated as a multi-class classification problem~\cite{lyu2021mdnn}.

Over the years, many computational approaches have been developed for drug-drug interaction predictions~\cite{lyu2021mdnn,wang2022predicting,Han2022,Xiong2023}. Earlier approaches were mostly based on traditional machine learning algorithms and/or matrix decomposition framework~\cite{Han2022}. Some methods~\cite{yang2021safedrug,yang2023molerec,Chen2023} can even take into considerations of patients' medical history and recommend drugs that are safe to use  for specific patients. More recently, more and more approaches based on deep learning models have drawn increasing attention, and have achieved better performance~\cite{Han2022}. 
In particular, with advancements in the development of GNNs, methods that adapt GNNs as backbones to incorporate interactions as networks have achieved state-of-the-art results~\cite{Chen2023}. 

\section{Representations of Molecules}
\begin{figure*}[!h]
\centering
\includegraphics[scale=0.4]{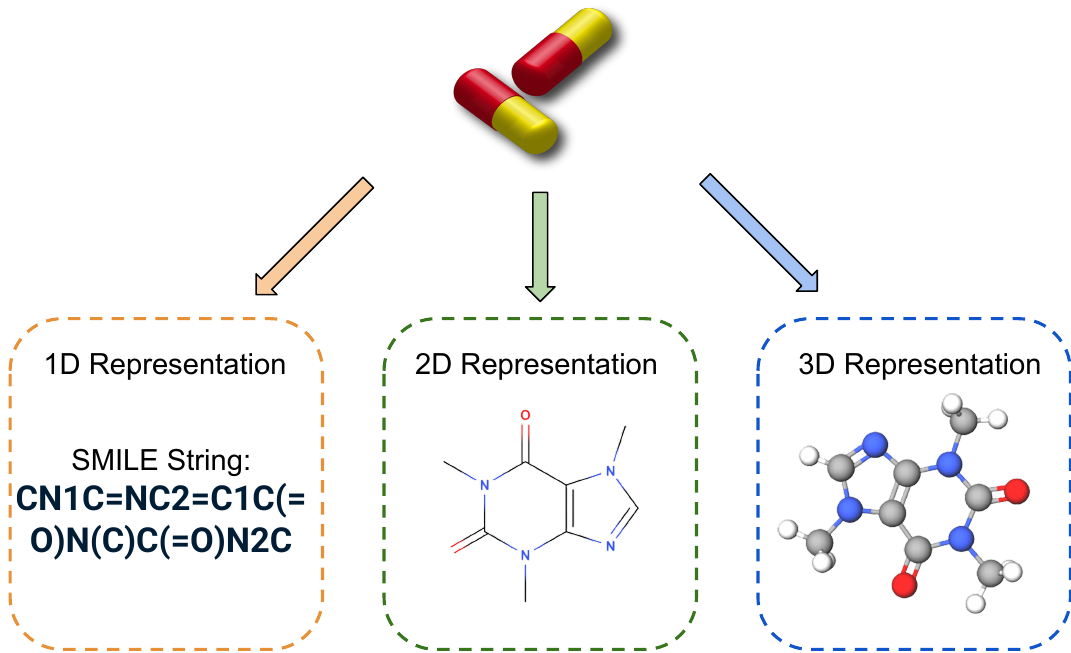}
\caption{Various molecular representations commonly employed in computational drug discovery models: 1D SMILES strings, 2D molecular graphs, and 3D molecular graphs.}
\label{fig:representation}
\end{figure*}

In general, molecules can be represented via fingerprints, the Simplified Molecular Input Line Entry System (SMILES) strings, or 2D-/3D-graphs in Fig.~\ref{fig:representation}. Binary fingerprints representing molecular substructure or topology allows efficient computation and database search~\cite{cereto2015molecular}. However, they cannot easily encode global features of molecules such as size and shape. SMILES is the most widely used linear representation for describing chemical structures since its invention~\cite{weininger1988smiles}, and is superior to other one-dimensional representation schemes such as binary fingerprints. However, there are innate limitations associated with the internal structure of SMILES representations when used in Natural Language Processing (NLP) algorithms.

Molecules can be represented as 2D graphs, where nodes represent atoms and edges represent chemical bonds, or as 3D graphs that also incorporate 3D coordinates, providing detailed spatial information. In both cases, both nodes and edges can have their own unique properties or features. While the 2D graph representation is simpler, 3D representation can better capture interactions based on distances and angles between atoms~\cite{liu2022spherical,jiao2023energy}, thus providing a more comprehensive view that is crucial for modeling molecular dynamics and bindings.

With molecules intuitively represented as graphs, GNNs offer a natural framework for handling and analyzing molecular data. This synergy has sparked extensive research, positioning GNNs at the forefront of innovation in molecular property and interaction prediction. GNNs allow nodes to aggregate information through their edges, creating comprehensive graph representations. Furthermore, by combining graph structures with neural networks, GNNs can readily handle both classification and regression tasks ~\cite{kipf2017semisupervised, chen2021structured,chen2022graph}. Therefore, many innovations in GNNs focus on graph representation learning, rather than specific prediction tasks. Most of the approaches to be discussed in this survey thus can be utilized in many downstream applications.

\section{Molecule Generation}
\begin{figure*}[!h]
\centering
\includegraphics[scale=0.35]{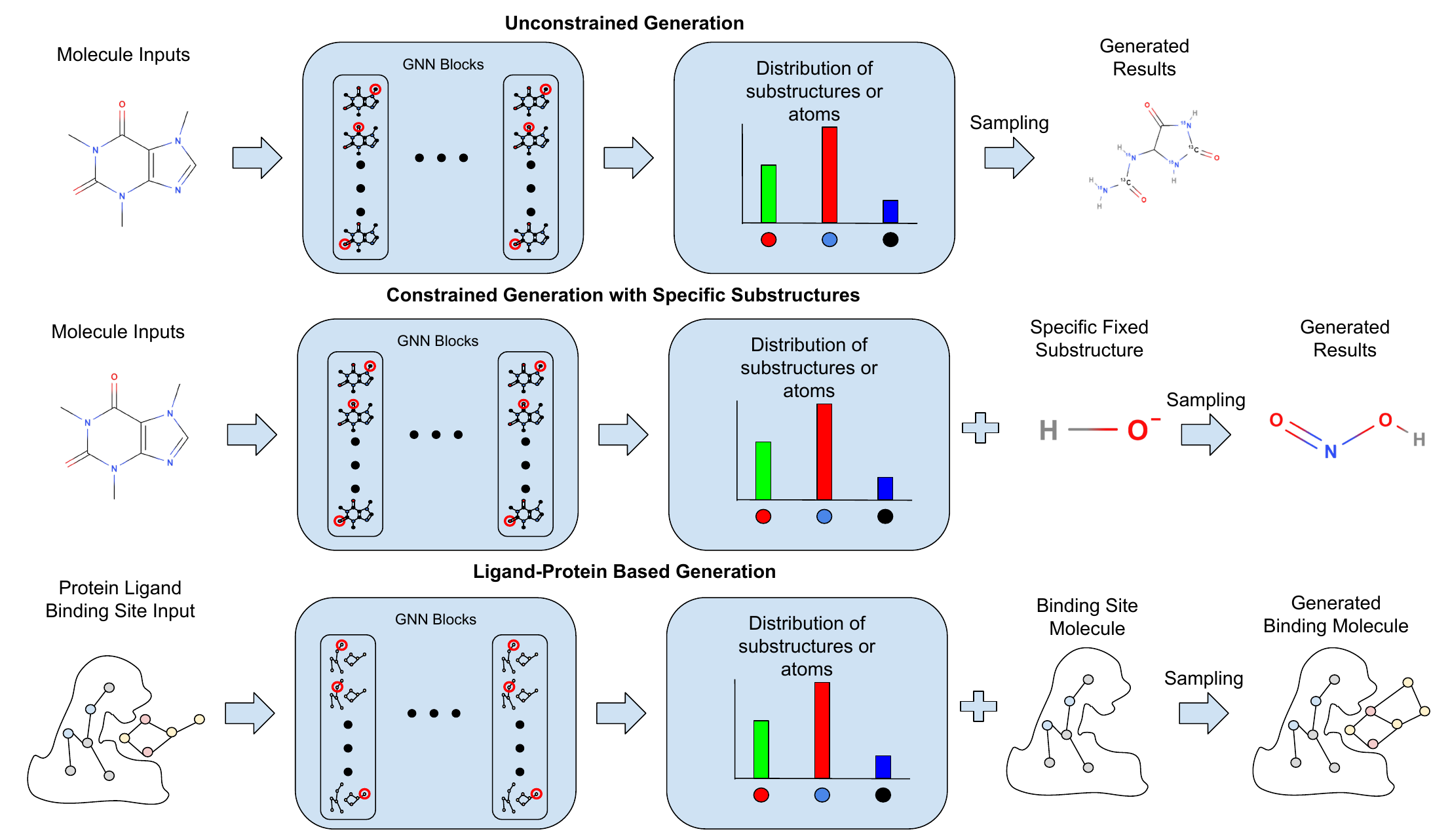}
\caption{The general framework of three different types of molecule generation processes. Molecular graphs and protein-ligand complexes are fed into GNN backbone models, which output the probability distribution of molecular substructures to be sampled, based on which the models select substructures and assemble the resulting molecules.}
\label{fig:generation}
\end{figure*}

One of the initial steps in drug discovery involves selecting a set of candidate molecules for early-stage screening. Traditionally, this process has been conducted in laboratories. However, recent advances in computational methods have expanded the toolkit available to researchers, including the use of GNN models for molecule generation. These models leverage 2D and 3D graph representations of molecules to efficiently generate novel compounds within defined constraints. The development of GNN-based molecule generation methods can be broadly categorized into three types, as illustrated in Fig.~\ref{fig:generation}: unconstrained generation~\cite{xu2021endtoend,swanson2023von,xu2023kernel,wang2024enhancing}, constrained generation with targeted substructures~\cite{maziarz2022learning,anonymous2024drug,geng2023novo}, and ligand-protein-based generation~\cite{luo20213d,liu2022generating,peng2022pocket2mol,zhang2023molecule,adams2022equivariant}.
Unconstrained methods prioritize structural diversity, constrained approaches focus on generating molecules containing specific functional groups or motifs relevant to desired chemical or biological properties, and ligand–protein-based strategies are designed to produce molecules that interact with specific protein targets. These advances demonstrate the versatility and promise of GNNs in accelerating drug discovery and highlight their pivotal role in identifying novel therapeutic candidates~\cite{Alves2022, Visan2024, Atance2022,xudiscrete}.

\subsection{Unconstrained Generation}
For computational methods to generate promising drug candidates, a natural approach is to generate chemically valid molecules in the vicinity of existing drugs and drug candidates (\textit{i.e.}, known training data). In other words, computational models are typically trained on ``hit molecules''--compounds that already exhibit desirable properties for drug development. Because there are no additional constraints on molecular substructures or chemical properties (\textit{e.g.}, binding affinity) imposed explicitly, such methods are classified as unconstrained. The primary goal of such approaches is to generate structurally valid and chemically plausible molecules with high diversity, guided by the distribution of the training data.

GraphINVENT~\cite{Mercado2021} is one of the initial explorations into unconstrained molecular graph generation, which explored various GNN architectures--such as gated GNNs and attention-based GNNs--to learn an ``action'' probability distribution from training molecules. The model then iteratively sampled one  action at a time, \textit{e.g.}, adding a bond or an atom, until a ``terminate generation'' action was reached. Doing so allowed the approach to construct a molecule step by step. GraphINVENT is a representative among a broader class of models~\cite{li2018learning, assouel2018defactor, you2018graphrnn, NEURIPS2019_46d0671d} that all adopt such an iterative sampling strategy that build new graphs by repeatedly sampling components or actions from learned distributions.

Beyond this one-piece-at-a-time paradigm, another type of method adopted generative frameworks that produce complete molecules (graphs) in a single pass. For instance, ConfVAE~\cite{xu2021end} integrated both 2D molecular graphs and 3D conformations, ensuring rotational and translational invariance. It employed a Conditional Variational Autoencoder (CVAE) framework, with Message Passing Neural Networks (MPNNs) for graph encoding, enabling end-to-end conditional molecule generation. While ConfVAE leveraged GNNs within a generative framework, VonMisesNet~\cite{swanson2023von} took a different approach by focusing on capturing the realistic distribution of torsional angles in molecules. It introduced a novel GNN architecture that sampled torsion angles from the Von Mises distribution, which better reflected the physical constraints of molecular geometry. Moreover, VonMisesNet addressed key challenges such as chirality inversion of atoms and supports molecules with a large number of rotatable bonds-enhancing the chemical accuracy and diversity of its outputs.

Overall, unconstrained generation models are increasingly integrating with generative architectures such as VAEs and GANs, aiming to better approximate the true data distribution. Nonetheless, as the need grows to reduce the cost of candidate selection, improve the quality of generated molecules, and generate molecules that could be synthesized,  research is gradually shifting toward generation models that incorporate explicit constraints on substructures and target properties.

\subsection{Constrained Generation with Specific Substructures} 
In drug development, generating molecules with specific substructures and targeted chemical properties is often more desirable than unconstrained molecule generation. MoLeR, introduced by Maziarz et al.~\cite{maziarz2022learning}, demonstrated the capability to perform both constrained and unconstrained molecule generation. It utilized motifs--common chemical substructures--within an encoder-decoder framework. By combining GNN and Multilayer Perceptron (MLP) blocks, MoLeR carefully constructed molecules one motif at a time, sequentially selecting motifs or atoms, determining attachment points, and assigning bond types, each step optimized for molecular validity and functionality.

Building on the success of fragment-based approaches like MoLeR, newer models have further advanced constrained molecule generation. One such model, GEAM~\cite{anonymous2024drug}, introduced the Graph Information Bottleneck (GIB) principle to identify substructures most relevant to specific drug properties. GEAM first extracted a vocabulary of meaningful substructures and then assembled molecules from this learned vocabulary. A soft actor-critic (SAC) reinforcement learning algorithm was used to identify high-quality samples, which were subsequently mutated through a genetic algorithm (GA) to produce final molecules that were chemically valid and aligned with the desired drug properties.

While models like MoLeR and GEAM focused primarily on the generation process, MiCam~\cite{geng2023novo} proposed a novel strategy for building a chemically ``reasonable" motif vocabulary. MiCam addressed the limitation of previous fragment-based methods, which often failed to identify appropriate motifs for molecule generation. Its vocabulary construction involved two phases: in the merging-operation learning phase, the model iteratively merged the most frequent atomic patterns found across molecules to form a preliminary set of motifs. In the motif-vocabulary construction phase, the model disconnected fragments at learned attachment points, marking these connection sites with special tokens to preserve the information necessary for molecule assembly. This approach allowed MiCam to flexibly generate molecules either by adding known motifs or by extending partially generated structures based on the connection history.

Overall, these models share a common strategy of using substructures as modular building blocks, aligning generation objectives with the training loss, and constructing molecules in a stepwise manner. Among the three, GEAM and MiCam offer greater flexibility in incorporating specific constraints on both substructures and chemical properties, as they allow the use of both atoms and motifs during generation. In contrast, MoLeR primarily relies on starting from a predefined scaffold.

\subsection{Protein-Ligand based Generation}
In addition to generating molecules based solely on training data of individual compounds, researchers have developed models that focus on protein binding sites and their associated ligands, addressing a new set of challenges in drug discovery. Recent advances in GNN-based molecule generation have enabled the creation of molecules specifically tailored for target proteins. These models employ GNN blocks to maintain structural consistency—ensuring robustness against flips, shifts, and rotations—while processing attributes and 3D coordinates of protein binding sites. Approaches such as the AR model~\cite{luo20213d} and GraphBP~\cite{liu2022generating} introduced distinct strategies for representing atoms and binding environments.

These models adopted various techniques to prioritize contextual representation and resilience to rigid transformations. For instance, AR combined MLP blocks with an auxiliary network to guide atom generation and bonding decisions, whereas GraphBP utilized spherical coordinates alongside MLPs for sequential atom-by-atom construction. Other notable methods, including Pocket2Mol~\cite{peng2022pocket2mol} and FLAG~\cite{zhang2023molecule}, incorporated auxiliary MLP classifiers and predictors to optimize atom positioning and motif attachment. Collectively, these strategies significantly improved model robustness and adaptability, representing critical progress toward customizing molecules for specific protein targets—a key advancement in drug discovery.

Each model has distinct strengths: the AR model emphasized specificity and binding affinity optimization for particular protein site structures; GraphBP introduced a dual-diffusion architecture to enhance flexibility; Pocket2Mol achieved greater computational efficiency through conditional 3D coordinate sampling; and FLAG leveraged motif-based generation to improve structural realism and diversity.

Beyond generating molecules that bind to specific protein sites, there is also a growing need to generate molecules constrained by a desired 3D binding conformation. In many cases, experimental data reveal that certain binding postures are particularly effective for interacting with specific proteins, and generating candidate molecules that adopt these conformations can greatly accelerate screening. SQUID~\cite{adams2022equivariant} was the first model designed to address this challenge. Given a target 3D shape, SQUID encodes the input conformation—treated as an unordered point cloud—into hidden features using GNN layers, and then iteratively generates 3D molecular fragments that reconstruct the desired shape fragment-by-fragment.

\begin{figure*}[!h]
\centering
\includegraphics[scale=0.28]{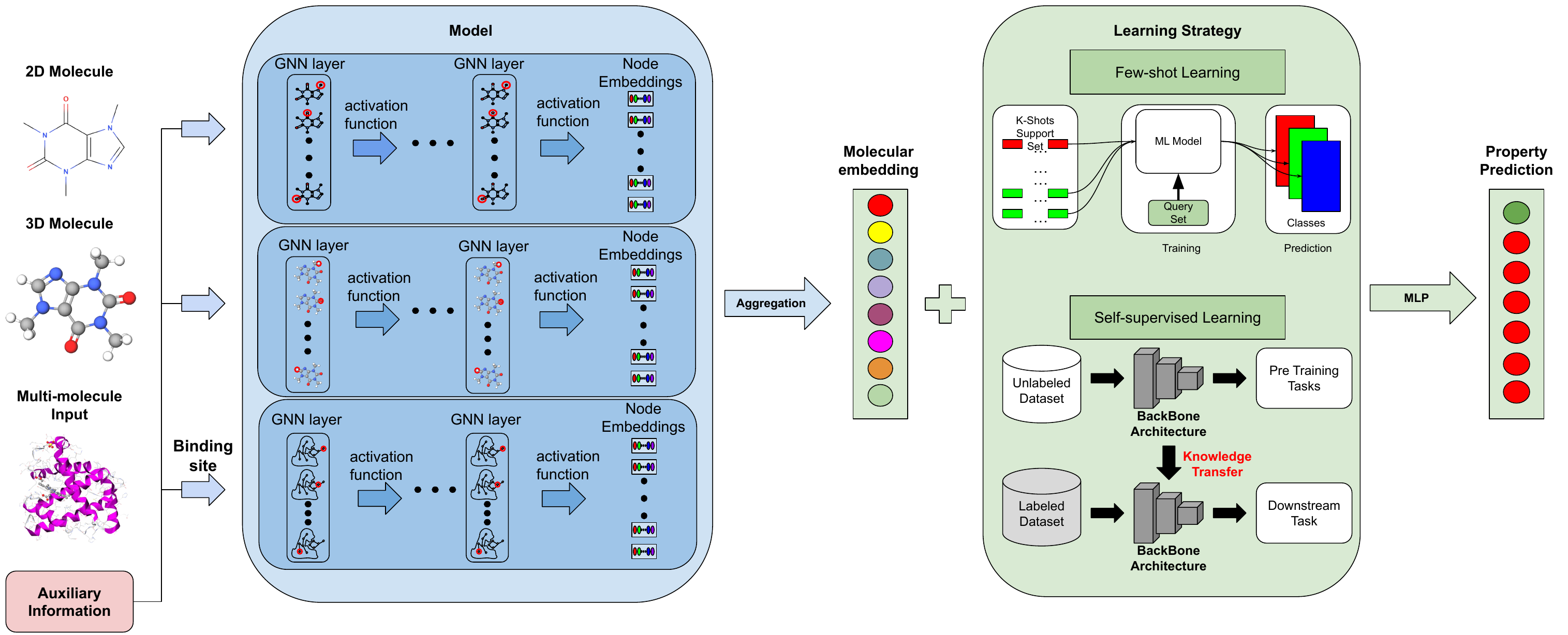}
\caption{The general framework for GNN-based drug property and interaction prediction. Three common types of inputs are used individually or jointly: 2D molecule graphs, 3D molecule graphs, multi-molecule interaction graphs such as protein-ligand complexes. Additional auxiliary information can also be incorporated by some approaches.  These inputs are then fed into GNN models, which aggregate information from neighboring nodes and produced final latent node representations. To alleviate the label sparsity issue,  various learning strategies, such as few-shot learning or self-supervised learning, are widely adopted.}
\label{fig:property}
\end{figure*}

\section{Prediction of Molecular Properties and Interactions}
In this section, we review various GNN-based approaches for predicting molecular properties and interactions. Different algorithms have been developed for different applications, each utilizing varying types of inputs. Common types of input data include 2D molecular graphs, 3D molecular conformations, multi-molecule complexes, and potentially additional auxiliary information (Fig.~\ref{fig:property}). 

A key component shared by all approaches is the learning of a latent molecular representation through GNNs, which is subsequently used for prediction tasks. To address challenges such as label sparsity, many methods also incorporate advanced learning strategies, including self-supervised (pre-training) and few-shot learning. In this section, we categorize and discuss these approaches based on their specific prediction tasks and input data types, highlighting their novel contributions and offering insights into how they advance the field.

\subsection{Property Prediction and Molecule Representation Learning Based on 2D Graphs}
Learning effective molecular representations is a fundamental step in property prediction. Generally, the goal of molecular representation learning is to embed molecules into numerical vectors in a latent space, enabling a variety of downstream tasks~\cite{chen2020learning}. Therefore, much of the innovation in property prediction from 2D graphs lies in the strategies developed for learning molecular representations, which is the focus of this subsection.

When molecules are represented as 2D graphs, early studies typically employed MPNNs~\cite{wieder2020compact}, wherein nodes (atoms) exchange messages with neighboring nodes, aggregating this information to update their respective states. However, obtaining class labels or quantitative measurements often requires costly laboratory experiments or human annotations. As a result, many datasets contain large numbers of unlabeled or imbalanced samples. To address this, newer approaches often incorporate pre-training strategies or rely on few-shot learning.

\subsubsection{Pre-training}
Pre-training\footnote{Here, pre-training refers to self-supervised learning on the input data itself or its augmented versions, differing slightly from the conventional notion in transfer learning where models are first trained on large independent datasets.} has emerged as a promising technique to mitigate label scarcity and is widely adopted in modern GNN models for molecular property prediction. For example, MGSSL~\cite{zhang2021motif} introduced a motif-based graph self-supervised learning framework that exploited rich information in subgraphs often overlooked at the node level. In this setting, the choice of molecular fragmentation method is critical, as poor fragmentation can lead to suboptimal motif generation and degraded model performance.

Many other methods adopt contrastive graph learning as their pre-training strategy. MoCL~\cite{sun2021mocl}, for instance, utilized knowledge-aware contrastive learning informed by local and global domain knowledge. Local domain knowledge ensured semantic invariance during augmentation, while global domain knowledge infused structural similarity across the learning process. Nevertheless, designing augmentation schemes that generalize across diverse molecular structures remains challenging. KCL~\cite{fang2022molecular} combined contrastive learning with domain-specific knowledge graphs, offering tailored augmentations at the cost of generalizability. MCHNN~\cite{liu2023multi} applied multi-view contrastive learning with task-specific augmentations, enhancing the expressiveness of molecular representations. HiMol~\cite{zang2023hierarchical} advanced self-supervised learning by proposing a hierarchical GNN that captured node, motif, and graph-level representations. However, constructing motif dictionaries can be computationally intensive, especially for large molecular databases.

\subsubsection{Few-shot Learning}
Few-shot learning approaches aim to predict molecular properties with minimal labeled data~\cite{wang2023federated}. HSL-RG~\cite{ju2023few} explored both global and local structural semantics for few-shot learning: global information was captured via molecular relation graphs built from graph kernels, while local information was learned through transformation-invariant representations. Similarly, MHNfs~\cite{schimunek2023context} employed a context module that retrieved and enriched molecule representations from a large pool of reference molecules, although the requirement for such large reference sets may limit its practical use.

GS-Meta~\cite{zhuang2023graph} extended few-shot learning to simultaneously handle multiple properties or labels. PACIA~\cite{anonymous2024pacia} introduced hypernetworks to generate adaptive parameters for modulating the GNN encoder, reducing overfitting while maintaining flexibility. However, designing effective hypernetworks demands significant domain expertise and may constrain the method’s generalizability across tasks. An alternative strategy to combat label scarcity involves grammar-based generation, as exemplified by Geo-DEG~\cite{guo2023hierarchical}, which employed a hierarchical molecular grammar to create molecular graphs, using production paths as informative priors for structural similarity.

\subsubsection{Incorporating Auxiliary Information}
Beyond the standard 2D graph representation, researchers have explored integrating additional molecular information during learning. For example, PhysChem~\cite{yang2021deep} showed that incorporating physical and chemical properties improved molecular representations, though its cooperation mechanism was less interpretable than traditional ensemble strategies. O-GNN~\cite{zhu2022mathcal} incorporated ring priors into the modeling, leveraging their importance in determining molecular properties. MoleOOD~\cite{yang2022learning} introduced invariant substructure learning to better handle distribution shifts across environments. However, in datasets with large environmental variations, MoleOOD’s advantage over simpler methods like ERM~\cite{vapnik2013nature} may diminish.

Some recent methods have also integrated both one-dimensional sequential encodings (\textit{e.g.}, SMILES strings) and 2D graphs to jointly leverage information from both views. DVMP~\cite{zhu2023dual}, for instance, encoded molecular graphs via GNNs and SMILES sequences via Transformers, using a dual-view consistency loss to maintain semantic coherence. However, involving both Transformer and GNN branches significantly increases training costs compared to single-branch models.

\subsection{Property Prediction based on 3D-graphs}

Recent advancements in deep learning and molecular science, driven by the increasing availability of large-scale 3D molecular datasets, have significantly advanced property prediction based on 3D molecular graphs. SphereNet~\cite{liu2022spherical} introduced an innovative approach for 3D molecular representation learning, proposing a spherical message-passing scheme that explicitly incorporates 3D spatial information. While SphereNet demonstrated strong predictive performance, it lacked transparency and interpretability, which hinders the understanding of its decision-making processes. MolKGNN~\cite{liu2023interpretable} addressed this limitation in the context of quantitative structure–activity relationship (QSAR) modeling. It enhanced 3D molecular representation learning by employing molecular graph convolution with learnable molecular kernels, effectively capturing chemical patterns. Importantly, MolKGNN incorporated molecular chirality, a critical aspect often neglected in previous models. However, the method emphasizes distinguishing specific molecular substructures, which may limit its generalizability across diverse chemical variations encountered in practical drug discovery scenarios.

Several studies have explored integrating both 2D and 3D information for property prediction. For instance, GraphMVP~\cite{liu2021pre} developed a 2D graph encoder enriched by discriminative 3D geometric information. It employed a self-supervised pre-training strategy that leveraged the correspondence and consistency between 2D topologies and 3D conformations. Similarly, 3D-Informax~\cite{stark20223d} proposed a transfer learning framework that pre-trained on molecules with both 2D and 3D data and then transferred the learned knowledge to molecules with only 2D structures. However, such approaches may risk overfitting, as evidenced by performance variability across different datasets.

UnifiedPML~\cite{zhu2022unified} further improved representation learning by jointly considering 2D and 3D information in its pre-training scheme. The framework employed three complementary tasks: reconstruction of masked atoms and coordinates, generation of 3D conformations conditioned on 2D graphs, and generation of 2D graphs conditioned on 3D conformations. GeomGCL~\cite{li2022geomgcl} adopted a dual-channel message-passing neural network to effectively capture both topological and geometric features of molecular graphs.

MoleculeSDE~\cite{liu2023group} unified 2D and 3D molecular representations by treating them as separate modalities in a multi-modal pre-training framework. 3D-PGT~\cite{wang2023automated} proposed a generative pre-training approach on 3D molecular graphs, which was subsequently fine-tuned on molecules lacking 3D structural data. It employed a multi-task learning strategy based on three geometric descriptors--bond lengths, bond angles, and dihedral angles--and used total molecular energy as an optimization target. While promising, the effectiveness of this framework remains to be validated on larger and more structurally complex molecules, as current evaluations have primarily focused on small molecules.

\subsection{Interaction Prediction}
Beyond property prediction, interaction prediction has been extensively explored, especially for drug-target or drug -disease interaction predictions. Many researchers (e.g., NeurTN~\cite{chen2020learning}) have utilized drug-target interaction networks as input, where the nodes are drugs and targets and links are known drug-target relationship. These models typically infer new interactions based on the guilt-by-association principle and are fundamentally different from the methods discussed in this work, which rely primarily on drugs' structure information represented as molecular graphs. Therefore, interaction network based approaches are excluded from further discussion in this section.

In drug discovery, one of the most critical and extensively studied relationships is the interaction between drugs (or chemical compounds) and their protein targets. In the literature, various terms are used to describe these interactions, each emphasizing different aspects. These include drug-target interaction, protein-ligand interaction, drug-target binding affinity, protein-ligand binding affinity, molecular docking. Computationally, given the 2D or 3D structures of two molecules, the interaction can be studied at three levels: (1) binary interaction (\textit{i.e.}, whether an interaction occurs), (2) binding affinity (a numerical value, typically reflecting binding free energy), and (3) docking or protein-ligand binding dynamics.

GNN-based approaches have been proposed to predict drug-target interactions based on their 2D structure. For example, CGIB~\cite{lee2023conditional} predicted interactions primarily using substructure information from paired graphs. MGraphDTA~\cite{yang2022mgraphdta} predicted drug–target binding affinities based on 2D compound graphs and protein sequences. It utilized a deep GNN to capture both local and global molecular structures and a multi-scale convolutional neural network (CNN) to extract features from protein sequences. However, capturing long-range dependencies within complex molecular graphs remains a challenge for such models. Given the superior performance generally observed when utilizing 3D molecular geometries, the trend shows that more approaches incorporate 3D information for interaction prediction, especially for binding affinity prediction and docking. 

For binding affinity prediction, the inputs are usually protein-ligand complexes, and the objective is to predict a binding score that reflects the strength of interaction, typically in terms of free energy. Recent developments have leveraged 3D graph representation learning to tackle this problem. For instance, Jones et al.~\cite{jones2021improved} proposed a fusion model that combined complementary molecular representations. Their method utilized a 3D CNN to capture local spatial features and a spatial GNN to encode global structural information, integrating both in a fused architecture.

The IGN framework~\cite{jiang2021interactiongraphnet} modeled protein-ligand complexes using three distinct molecular graphs, each incorporating both 3D structural and chemical properties. MP-GNN~\cite{li2022multiphysical} introduced a multiphysical molecular graph representation, which systematically captured a wide range of molecular interactions across different atom types and physical scales. However, most existing biomolecular GNNs rely on covalent-bond-based graph constructions, which often fail to effectively characterize non-covalent interactions essential for modeling biomolecular complexes. 

GraphscoreDTA~\cite{wang2023graphscoredta} advanced this field by integrating a bitransport information mechanism and Vina distance optimization terms to better capture the mutual information between proteins and ligands. This method also highlighted critical atomic and residue-level features. In contrast to the above, NERE~\cite{jin2023unsupervised} proposed an unsupervised approach to binding energy prediction, framing it as a generative modeling task. Their method, based on Neural Euler’s Rotation Equations (NERE), predicted molecular rotations by modeling the forces and torques between ligand and protein atoms. However, the current implementation of NERE for antibody modeling only considers backbone atoms and omits side-chain atoms, which are crucial for accurately estimating binding affinity.

Docking, a central process in drug discovery, has also seen innovation through GNN-based approaches. E3Bind~\cite{zhang2022e3bind} introduced an end-to-end model that directly generates ligand coordinates, thus eliminating the need for traditional sampling procedures and coordinate reconstructions. Similarly, FABind~\cite{pei2023fabind} combined pocket prediction and docking in an integrated model for fast and accurate binding pose prediction. A unique ligand-informed pocket prediction module was used to guide the docking process, with successive refinements optimizing the ligand-protein binding pose. The model further enhanced the docking process by incrementally integrating the predicted pockets to optimize protein-ligand binding. However, ablation studies indicated that different components contribute to the model's performance in varying degrees, suggesting potential inefficiencies in the overall architecture. More recently, NeuralMD~\cite{liu2024multigrained} provided a fine-grained simulation of protein-ligand binding dynamics. The model included BindingNet, which adhered to group symmetry and captured multi-level interactions, and a neural ordinary differential equation (ODE) solver that modeled the physical trajectories of atoms based on Newtonian mechanics.

EquiPocket~\cite{zhang2023equipocket}, distinct from the aforementioned methods, focused specifically on predicting ligand binding sites for given 3D protein structures. It introduced three novel modules: a local geometric modeling module to extract features from individual surface atoms, a global structural module to encode the chemical and spatial context of the entire protein, and a surface message-passing module to learn surface-level geometric patterns. In contrast to CNN-based methods, which suffer from inefficiencies due to voxelization of irregular protein surfaces, EquiPocket avoids computational redundancy and excessive memory usage through its surface-based geometric design.

\begin{figure*}[!h]
\centering
\includegraphics[scale=0.35]{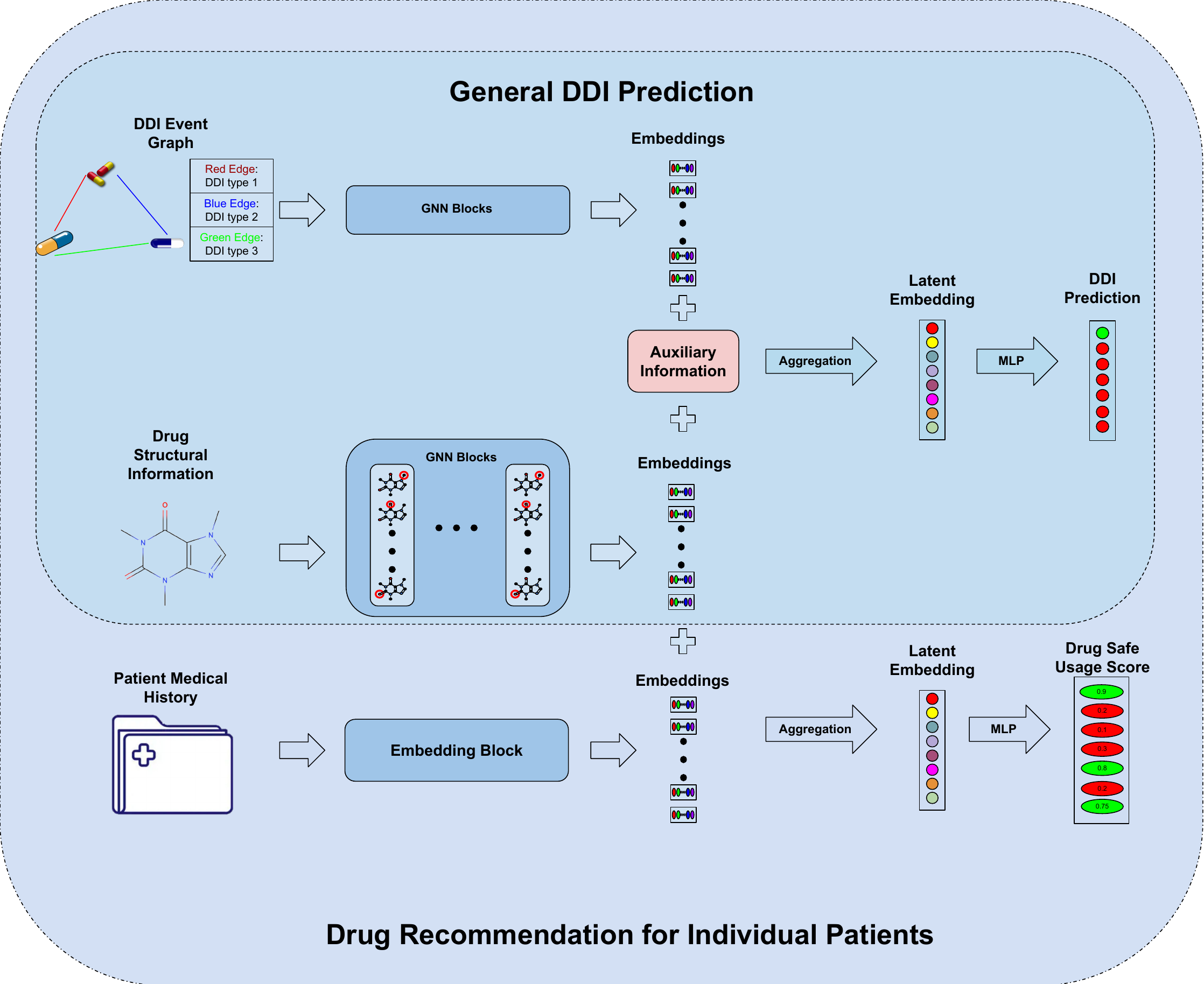}
\caption{The general process of DDI prediction based on  GNN models. Possible inputs for the general DDI prediction (the one inside the small rectangle) include DDI event graphs, and drug molecular structures, either individually or jointly. Additional auxiliary data can be incorporated into the models. GNN blocks map the inputs into the latent space, which will be utilized for DDI prediction. By including patient medical history, the model can be extended to perform patient-specific drug safety recommendations.}
\label{fig:ddi}
\end{figure*}

\section{Prediction of Drug-Drug Interactions}

Predicting and understanding DDIs is a critical step in computational drug discovery, especially in the context of drug combination therapies~\cite{Chou2006, Lehr2009, Liu2025}, in which case multiple drugs are commonly used together in clinical practice  to treat complex diseases such as cancer~\cite{Borisy2003, Lee2007, Han2017}. However, polypharmacy elevates the risk of adverse DDIs, potentially compromising therapeutic efficacy, posing serious health risks, and increasing healthcare costs~\cite{Pirmohamed2004, Dechanont2014, Juurlink2003}. 
Historically, many DDIs were discovered via clinical case reports or mined from electronic health records (EHRs)~\cite{Iyer2014,Wu2021}.
Computational approaches, particularly those based on machine learning, now offer scalable and cost-effective alternatives to identify novel candidate interactions, either synergistic or adverse ones, beforehand.

Recent computational strategies for DDI prediction can be broadly categorized into two paradigms, illustrated in Fig.~\ref{fig:ddi}: (1) \emph{general DDI prediction}, which identifies potential interactions across large drug populations; and (2) \emph{personalized drug combination recommendation}, which tailors treatment regimens by further considering individual patient health profiles and personal DDI risk. In this subsection, we will first focus on general DDI prediction, and examine commonly used input data types, followed by the discussion of recent developments in problem formulations and model architectures. We conclude this subsection with discussions on personalized drug combination recommendation. 

\subsection{Types of Input Data}
Similar to drug–target interaction prediction discussed in the previous section, many early approaches to DDI prediction primarily utilized drug–drug interaction graphs, where nodes represent drugs and edges encode their interactions. With advancements in the field, more recent methods have begun to incorporate drug molecular structures, represented as molecular graphs (as introduced previously), in which nodes denote atoms and edges correspond to chemical bonds. In both types of graphs, nodes and edges are typically enriched with additional features or attributes that capture relevant properties. For example, in drug–drug interaction graphs, node attributes may encode drug-specific properties, while edges can be labeled to indicate interaction types (e.g., synergistic or antagonistic effects). In molecular graphs, node features represent atomic properties, and edge features describe bond characteristics. GNN-based models are capable of processing both graph types, often alongside auxiliary information such as drug similarity matrices, to learn latent representations from different perspectives and at multiple levels—capturing both relational patterns in drug interaction networks and structural characteristics at molecular and sub-molecular scales. Increasingly, recent approaches aim to integrate both types of input within a unified learning framework, jointly capturing topological  and structural information to enhance predictive performance.

\subsection{Problem Formulations}
The choice of input data and their representations not only inspires model design but also significantly influences problem formulations. Early models primarily focused on drug–drug interaction graphs combined with simple node-level drug features. Various GNN architectures were employed to learn low-dimensional representations of drugs from these graphs~\cite{huang2020skipgnn,feng2020dpddi}. Building on this idea, models such as GCNMK~\cite{wang2022predicting} further decomposed the DDI graph into two separate graphs, one representing interactions where a drug increases the activity of another, and the other where it decreases activity, and applied two GNNs to learn drug representations from these differentiated views. Subsequent studies expanded the input space by incorporating drug molecular graphs, thereby enabling the integration of both structural and relational perspectives to enhance model performance. For instance, MRCGNN~\cite{Xiong2023} employed a GNN to process the relational DDI graph while enriching each drug's feature representation with molecular-level information extracted by a separate GNN operating on its molecular graph. This multimodal approach allowed the model to simultaneously capture both chemical and interaction-level knowledge.

The evolution of problem formulation also extends to the design of prediction tasks. Some models framed DDI prediction as a binary classification problem, aiming solely to determine the existence of an interaction~\cite{feng2020dpddi}. Others formulated it as a multi-label classification task, predicting both the presence and the specific type of interaction from a predefined label set~\cite{Xu_2019, bai2020, Zhong2023, Gao2024}. Many researchers have further distinguished between adverse DDI prediction~\cite{Zitnik2018, Deac2019, Liu2022} and drug combination or synergy prediction~\cite{chen2019adversarial, Liu2022, Wang2021, Zhang2023}, providing a more nuanced understanding of interaction consequences.

Overall, advancements in problem formulation aim to enrich input representations with biologically meaningful information and to enable more fine-grained, application-specific predictions. Future directions are likely to emphasize input data that better reflect the underlying biological complexity of DDIs. For the prediction outcomes, knowledge of specific side effects from adverse interaction prediction and information of targeted diseases in drug combination modeling are welcome additions.

\subsection{Advancements in Model Structure}

Beyond problem formulation, significant research has focused on improving model architectures to more effectively aggregate information across different data modalities and drug representations. These approaches typically employ distinct architectures (e.g., GNNs, CNNs) for different modalities or representations, and combine their learned features using various fusion strategies. Earlier models such as MDNN~\cite{lyu2021mdnn}, GCNMK~\cite{wang2022predicting}, MRCGNN~\cite{Xiong2023}, and DeepDDS~\cite{Wang2021} fused features from each modality or representation through simple concatenation. While this strategy preserves feature information from all modalities, it neither accounts for the relative importance of each data source nor captures potential inter-modal relationships.

To address these limitations, more recent models have incorporated attention mechanisms to fuse latent features from different modalities, or from different drugs within the same modality, via cross-attention. For instance, SSF-DDI~\cite{Zhu2024} utilized two drug representations: the 1D SMILES sequence and the 2D graph structure. Separate architectures (CNN for SMILES and GNN for molecular graphs) were used, and a cross-attention mechanism was employed to integrate the latent features generated by the two models. Similarly, SRR-DDI~\cite{NIU2024111337} constructed 2D molecular graph representations for drug pairs and applied cross-attention to fuse the learned latent features of the two drugs.

MD-Syn~\cite{ge2025} proposed a multi-modal architecture with both one-dimensional and two-dimensional feature embedding modules, which allows incorporation of SMILES sequences, cell line information, drug molecular graphs, and protein–protein interaction (PPI) networks. Rather than using cross-attention, MD-Syn introduced a graph-trans pooling module within the 2D-feature embedding module, employing Transformer encoder layers with multi-head self-attention to process the concatenated latent representations from the PPI network and drug graphs.

Another direction in architectural advancement focuses on multi-level feature aggregation across GNN layers, particularly for molecular graphs. For example, DAS-DDI~\cite{NIU2024104672} introduced weighted layer-wise aggregation, where each GNN layer contributes differently to the final embedding. This enables molecular substructures of varying granularity to inform the final drug representation, thereby enhancing the expressiveness and robustness of the model in capturing complex inter-drug relationships.

\begin{table*}[h!]
\small
\centering
\caption{GNN-based models discussed in this review and their characteristics. Each row includes the name of the approach, the main model architecture, the prediction task and the datasets used. The approaches are grouped into different bucket based on their tasks. The background with yellow color indicates that the approaches primarily utilized 2D structure and the blue color indicates that the approaches primarily utilized 3D structure. Methods using pre-training are labeled with $\sharp$ and methods using few-shot learning are labeled with $\star$.}
\label{tab:methods}
\resizebox{1.0\textwidth}{!}{%
\begin{tabular}{l|l|l|l}
\toprule
\textbf{Name} & \textbf{architecture} & \textbf{Task} & \textbf{Datasets}\\
\midrule
\rowcolor{babyblue} ConfVAE~\cite{xu2021endtoend} & MPNN &Unconstrained Generation w/ CVAE and uses 2D\&3D& GEOM-QM9, GEOM-Drugs \\
\rowcolor{babyblue} VonMisesNet~\cite{swanson2023von} & GCN &Unconstrained Generation w/ Von Mises distribution& NMRShiftDB, GDB-17 \\
\midrule
\rowcolor{babyblue} MoLeR~\cite{maziarz2022learning} & GNN &Constrained Generation w/ motifs-based substructures& GuacaMol \\
\rowcolor{babyblue} MiCam~\cite{geng2023novo} & GNN &Constrained Generation w/ connection-aware motif vocabulary& QM9, ZINC, GuacaMol \\
\rowcolor{babyblue} GEAM~\cite{anonymous2024drug} & MPNN &Constrained Generation w/ soft-actor critic& ZINC250k \\
\midrule
\rowcolor{babyblue} AR~\cite{luo20213d} & GNN &Ligand-Protein Based Generation w/ auxiliary network& CrossDocked \\
\rowcolor{babyblue} GraphBP~\cite{liu2022generating} & GNN  &Ligand-Protein Based Generation w/ spherical coordinates& CrossDocked \\
\rowcolor{babyblue} Pocket2Mol~\cite{peng2022pocket2mol} & GNN &Ligand-Protein Based Generation w/ auxiliary atom positioning& CrossDocked \\
\rowcolor{babyblue} FLAG~\cite{zhang2023molecule} & GNN &Ligand-Protein Based Generation w/ auxiliary motif attachment&  CrossDocked \\
\rowcolor{babyblue} SQUID~\cite{adams2022equivariant} & GNN &Ligand-Protein Based Generation w/ 3-D shape&  MOSES \\
\midrule
\rowcolor{bananamania} NeurTN~\cite{chen2020learning} & GNN &Property Prediction w/ powerful nonlinear relationships& CTD,DrugBank,UniProt4\\
\rowcolor{bananamania} PhysChem~\cite{yang2021deep} & MPNN &Property Prediction w/ physical\&chemical information& QM7,QM8,QM9,Lipop,FreeSolv,ESOL,COVID19\\
\rowcolor{bananamania} O-GNN~\cite{zhu2022mathcal} & GNN &Property Prediction w/ ring substructures& BBBP,Tox21,ClinTox,HIV,BACE,SIDER,FS-Mol\\
\rowcolor{bananamania} MoleOOD~\cite{yang2022learning} & SAGE &Property Prediction w/ invariant substructure across environments& BACE,BBBP,SIDER,HIV,DrugOOD\\
\rowcolor{bananamania} MGSSL~\cite{zhang2021motif} & GNN $\sharp$&Property Prediction w/ motif-based self-supervised learning& MUV,ClinTox,SIDER,HIV,Tox21,BACE,\\
\rowcolor{bananamania} & & & ToxCast,BBBP\\
\rowcolor{bananamania} MoCL~\cite{sun2021mocl} & GIN $\sharp$&Property Prediction w/ knowledge-aware contrastive learning& BACE,BBBP,ClinTox,Mutag,SIDER,Tox21,ToxCast\\
\rowcolor{bananamania} KCL~\cite{fang2022molecular} & MPNN $\sharp$&Property Prediction w/ domain knowledge contrastive learning&BBBP,Tox21,ToxCast,SIDER,ClinTox,BACE,\\
\rowcolor{bananamania} & &  & ESOL,FreeSolv\\
\rowcolor{bananamania} MCHNN~\cite{liu2023multi} & GCN $\sharp$&Property Prediction w/ multi-view contrastive learning& PubChem,MDAD,DrugVirus,HMDAD,Disbiome,\\
\rowcolor{bananamania} & &  & gutMDisorder,Peryton\\
\rowcolor{bananamania} HiMol~\cite{zang2023hierarchical} & GNN $\sharp$& Property Prediction w/ boundaries self-supervised learning& BACE,BBBP,Tox21,ToxCast,SIDER,ClinTox,\\
\rowcolor{bananamania} & &  & ESOL,FreeSolv,Lipop,QM7,QM8,QM9\\
\rowcolor{bananamania} HSL-RG~\cite{ju2023few} & GNN $\sharp \star$ &Property Prediction w/ few-shot learning\&self-supervised learning& Tox21,SIDER,MUV,ToxCast\\
\rowcolor{bananamania} MHNfs~\cite{schimunek2023context} & GNN $\star$& Property Prediction w/ few-shot learning\&context module & FS-Mol\\
\rowcolor{bananamania} GS-Meta~\cite{zhuang2023graph} & GNN $\star$& Property Prediction w/ few-shot learning\&simultaneous multiple labels& Tox21,SIDER,MUV,ToxCast,PCBA\\
\rowcolor{bananamania} PACIA~\cite{anonymous2024pacia} & GNN $\sharp \star$&Property Prediction w/ few-shot learning\&adaptive parameters& Tox21,SIDER,MUV,ToxCast,FS-Mol\\
\rowcolor{bananamania} Geo-DEG~\cite{guo2023hierarchical} & MPNN & Property Prediction w/ hierarchical molecular grammar& CROW,Permeability,FreeSolv,Lipop,HOPV,PTC,ClinTox\\
\rowcolor{bananamania} DVMP~\cite{zhu2023dual} & GCN $\sharp$&Property Prediction w/ pre-train for dual-view 1D\&2D molecule& BBBP,Tox21,ClinTox,HIV,BACE,SIDER,ESOL\\
\midrule
\rowcolor{babyblue} GraphMVP~\cite{liu2021pre} & GNN $\sharp$&Property Prediction w/ pre-train consistency between 2D\&3D& BBBP,Tox21,ToxCast,SIDER,MUV,HIV,BACE\\
\rowcolor{babyblue} SphereNet~\cite{liu2022spherical} & MPNN &Property Prediction w/ spherical message passing& QM9 \\
\rowcolor{babyblue} UnifiedPML~\cite{zhu2022unified} & GN Blocks $\sharp$& Property Prediction w/ pre-train on multi-tasks for 2D\&3D& BBBP,Tox21,ClinTox,HIV,BACE,SIDER\\
\rowcolor{babyblue} GeomGCL~\cite{li2022geomgcl} & MPNN $\sharp$&Property Prediction w/ dual-channel message passing for 2D\&3D& ClinTox,SIDER,Tox21,ToxCast,ESOL,FreeSolv,Lipop\\
\rowcolor{babyblue} MolKGNN~\cite{liu2023interpretable} & GNN &Property Prediction w/ molecular chirality& PubChem\\
\rowcolor{babyblue} 3D-Informax~\cite{stark20223d} & MPNN $\sharp$&Property Prediction w/ transfer learning for 2D\&3D& QM9,GEOM-Drugs\\
\rowcolor{babyblue} MoleculeSDE~\cite{liu2023group} & GNN $\sharp$&Property Prediction w/ multi-modal pre-train for 2D\&3D& BBBP,Tox21,ToxCast,SIDER,ClinTox,MUV,HIV,BACE\\
\rowcolor{babyblue} 3D-PGT~\cite{wang2023automated} & GNN $\sharp$&Property Prediction w/ multi-task generative pre-train on 3D& BBBP,Tox21,ToxCast,SIDER,ClinTox,MUV,\\
\rowcolor{babyblue} &  & & HIV,BACE,ESOL,Lipop,Malaria,CEP,Davis,KIBA\\
\midrule
\rowcolor{bananamania} MGraphDTA~\cite{yang2022mgraphdta} & GNN &Molecular Interactions Prediction w/ super-deep GNN& Davis,KIBA,Metz,Human,C. elegans,ToxCast\\
\rowcolor{bananamania} CGIB~\cite{lee2023conditional} & MPNN & Molecular Interactions Prediction w/ substructure information& MNSol,FreeSolv,CompSol,Abraham,CombiSolv\\
\midrule
\rowcolor{babyblue} SG-CNN~\cite{jones2021improved} & GNN & Binding Affinity Prediction w/ complementary representations& PDBbind \\
\rowcolor{babyblue} IGN~\cite{jiang2021interactiongraphnet} & GNN & Binding Affinity Prediction w/ chemical information& PDBbind\\
\rowcolor{babyblue} MP-GNN~\cite{li2022multiphysical} & GNN & Binding Affinity Prediction w/ multiphysical representations& PDBbind,SARS-CoV BA\\
\rowcolor{babyblue} GraphscoreDTA~\cite{wang2023graphscoredta} & GNN & Binding Affinity Prediction w/ bitransport information&PDBbind \\
\rowcolor{babyblue} NERE~\cite{jin2023unsupervised} & MPNN $\sharp$& Binding Affinity Prediction w/ Neural Euler's Rotation Equations& PDBbind\\
\rowcolor{babyblue} E3Bind~\cite{zhang2022e3bind} & GIN & Binding Affinity Prediction w/ docking& PDBbind\\
\rowcolor{babyblue} FABind~\cite{pei2023fabind} & GCN & Binding Affinity Prediction w/ pocket prediction and docking& PDBbind\\
\rowcolor{babyblue} NeuralMD~\cite{liu2024multigrained} & MPNN & Protein-Ligand Binding Dynamics Simulations& MISATO\\
\rowcolor{babyblue} EquiPocket~\cite{zhang2023equipocket} & GNN & Ligand Binding Site Prediction w/ geometric and chemical& scPDB,PDBbind,COACH420,HOLO4K\\
\midrule
\rowcolor{bananamania} MDNN~\cite{lyu2021mdnn} & GNN & DDI Prediction w/ knowledge graphs & DrugBank\\
\rowcolor{bananamania} DPDDI ~\cite{feng2020dpddi} & GCN &  DDI prediction
w/ extraction of the network structure features of drugs from DDI network &  DrugBank,ZhangDDI\\
\rowcolor{bananamania} GCNMK~\cite{wang2022predicting} & GCN & DDI Prediction w/ dual-block GNN & DrugBank\\
\rowcolor{bananamania} MRCGNN~\cite{Xiong2023} & GCN & DDI Prediction w/ incorporation of negative DDI event & Deng’s dataset,Ryu’s dataset\\
\rowcolor{bananamania} SRR-DDI~\cite{NIU2024111337} & MPNN & DDI Prediction w/ self-attention mechanism & DrugBank,Twosides\\
\rowcolor{bananamania} DAS-DDI~\cite{NIU2024104672} & GCN & DDI Prediction w/ dual-view framework & DrugBank,ChChMiner,ZhangDDI\\
\rowcolor{bananamania} SSF-DDI~\cite{Zhu2024} & MPNN & DDI Prediction w/
on sequence and substructure features & DrugBank,Twosides\\
\rowcolor{bananamania} DeepDDS~\cite{Wang2021} & GAT, GCN & synergetic DDI Prediction w/ attention mechanism & O’Neil’s dataset,Menden's dataset\\
\rowcolor{bananamania} MD-Syn~\cite{ge2025} & GCN &  synergistic DDI Prediction w/ chemicals and cancer cell line gene expression profiles & O’Neil’s dataset, DrugCombDB\\
\midrule
\rowcolor{bananamania} SafeDrug~\cite{yang2021safedrug} & MPNN & Drug Combinations Recommendation w/ explicit leverages of drugs’ molecule structures and model DDIs
 & MIMIC-III\\
\rowcolor{bananamania} MoleRec~\cite{yang2023molerec} & GIN &Drug Combinations Recommendation w/ molecular substructure-aware encoding method & MIMIC-III\\
\rowcolor{bananamania} Carmen~\cite{Chen2023} & GNN &Drug Combinations Recommendation w/ context-aware GNN& MIMIC-III,MIMIC-IV\\
\multicolumn{4}{l}{Graph Isomorphism Network(GIN), GraphSAGE(SAGE), Graph Convolutional Network(GCN), Graph network block(GN blocks)}\\
\bottomrule
\end{tabular}
}
\end{table*}

\subsection{Personalized Drug Combination Recommendation}

A distinct line of research focuses on personalized DDI prediction by incorporating patient-specific medical histories. These models are less common due to the fact that data privacy concerns hinder the availability of clinical data, but they offer unique insights. For instance, SafeDrug~\cite{yang2021safedrug} used GNNs and RNNs to align molecular features with patient treatment histories, producing compatibility scores for candidate drug combinations. MoleRec~\cite{yang2023molerec} leveraged attention mechanisms to integrate patient records and drug representations for safe prescription generation.

Despite their promise, challenges remain. GNNs often generate nearly identical embeddings for structurally similar molecules, regardless of therapeutic context. Carmen~\cite{Chen2023} addressed this with a context-aware GNN that incorporated medication context during atom-level message aggregation. This architecture produced distinct embeddings based on therapeutic relevance, offering a refined strategy for personalized drug combination recommendations. These models incorporating personal information represent a significant step toward safer, more effective treatment planning, highlighting the value of integrating biomedical knowledge with patient-specific data.

Finally, for easy reference, all the approaches discussed in this review for all the three tasks are organized in Table~\ref{tab:methods}.  


\section{Benchmark Databases}
In addition to newly developed methodologies, benchmark datasets play a vital role in advancing the field of computational drug discovery. High-quality data is essential for all the tasks ranging from molecular design and property prediction to the characterization of drug–drug interactions and it serves as a foundation for objectively evaluating the effectiveness of various predictive models. 

We assembled a comprehensive list of datasets referenced across the reviewed studies and organized them by their data characteristics, as summarized in Table~\ref {tab:datasets}. Four primary categories capture the scope of these resources: Comprehensive Databases, Clinical Databases, Structural Information Databases, and Molecular Interaction Databases. Given the breadth and interrelated nature of the latter, we subdivided Molecular Interaction Databases into Protein–Ligand Binding and Drug–Drug Interaction collections, each distinguished by color coding in table. While not exhaustive, this selection emphasizes the most influential datasets driving progress in computational drug discovery.

\begin{table*}[h!]
\small
\centering
\caption{Commonly used benchmark databases and their brief descriptions. Consistent with discussion in the paper, we separate the datasets into four categories: Comprehensive Databases, Clinical Databases, Structural Information Databases, and Molecular Interaction Databases. The Molecular Interaction category is further divided into Protein–Ligand Binding and Drug–Drug Interaction.}
\label{tab:datasets}
\resizebox{1.0\textwidth}{!}{
\begin{tabular}{l|l|l}
\toprule
\textbf{Task} & \textbf{Dataset} &  \textbf{Description} \\
\midrule

Comprehensive Databases & DrugBank~\cite{wishart2023drugbank6} & Extensive repository of approved and investigational drugs linking chemical structures with pharmacological profiles and target interactions. \\
& PubChem~\cite{kim2025pubchem} & Vast compound library annotated with high-throughput screening bioactivities and comprehensive chemical properties. \\
& MoleculeNet~\cite{wu2018moleculenet} & Aggregated benchmark collection covering diverse molecular properties and activities for algorithm evaluation. \\

\midrule

Clinical Databases & MIMIC-III~\cite{johnson2016mimic} & Detailed, de-identified ICU patient records including vitals, labs, and clinical interventions over time. \\
 & MIMIC-IV~\cite{Johnson2023} & A public EHR dataset with deidentified clinical data for 180,733 hospital and 50,920 ICU patients, covering patient tracking, billing, medications, and measurements. \\
& UK Biobank~\cite{ukbiobank} & Population-scale cohort with deep phenotypic, genotypic, and long-term health outcome data. \\

\midrule

Structural Information Databases & ZINC~\cite{irwin2020zinc22} & Vendor-curated set of purchasable compounds each with experimentally determined 3D conformers. \\
& GEOM~\cite{axelrod2022geom} & High-precision quantum-mechanically optimized 3D molecular geometries for conformational analysis. \\
& MISATO~\cite{siebenmorgen2024misato} & Multigrained collection of protein–ligand complexes annotated with binding-site details. \\
& CrossDocked~\cite{francoeur2020three} & Large-scale docking dataset providing multiple poses and affinity estimates for protein–ligand pairs. \\

\midrule

Molecular Interaction Databases & ChEMBL~\cite{zdrazil2024chembl} & Expert-curated database of small molecules linked to experimentally measured target binding affinities. \\
Protein–Ligand Binding & Metz Dataset~\cite{metz2011navigating} & Collection of kinase inhibitor experiments reporting inhibition constants (K\textsubscript{i}) across targets. \\
& KIBA Dataset~\cite{tang2014making} & Unified resource converting heterogeneous kinase-inhibitor bioactivities into standardized KIBA scores. \\
& Davis Dataset~\cite{davis2011comprehensive} & Comprehensive mapping of kinase–inhibitor dissociation constants (K\textsubscript{d}) over multiple enzymes. \\
& PDBbind Dataset~\cite{liu2017pdbbind} & Annotated set of biomolecular complexes with experimentally determined binding affinities and structures. \\

\midrule

Molecular Interaction Databases& TwoSIDES~\cite{tatonetti2012data} & Pharmacovigilance resource of adverse drug–drug event pairs mined from FAERS reporting data. \\
Drug–Drug Interaction& Deng’s Dataset~\cite{deng2020multimodal} & Multimodal catalog of 570 approved drugs’ interactions stratified by 65 mechanistic event types. \\
& ChChMiner~\cite{biosnapnets} & A BioSNAP sub-dataset of 1,514 FDA-approved drugs and 48,514 DDI. \\
& DrugCombDB~\cite{Liu2019} & Dataset that contains 448,555 combinations of 2,887 drugs across 124 cancer cell lines, labeled as synergistic or antagonistic using multiple scoring models. \\
& O'Neil's dataset~\cite{ONeil2016} & A dataset that contains 583 drug combinations across 39 cancer cell lines, identifying 287 synergistic and 178 antagonistic pairs among 38 drugs \\
& AstraZeneca's dataset~\cite{menden_community_2019} & A dataset that features 910 combinations of 118 drugs across 85 cell lines, with 797 pairs showing high synergy \\

\bottomrule
\end{tabular}}
\end{table*}

\subsection{Comprehensive Databases}
Comprehensive databases are those that contain extensive molecular and biochemical information on drugs and chemical compounds, including but not limited to compound identifiers, structural representations (e.g., SMILES, 2D and/or 3D graphs), indications, and target information. Such data supports a wide range of applications. The compound structure information usually serves as input for computational models and their labels and properties serve as training data. In this subsection, we include three representative comprehensive databases in drug discovery: DrugBank~\cite{wishart2023drugbank6}, PubChem~\cite{kim2025pubchem}, and MoleculeNet~\cite{wu2018moleculenet}. 

\textbf{DrugBank}~\cite{wishart2023drugbank6}  is a comprehensive, freely accessible, online database containing reliable information on drugs and drug target and is a vital resource for computational drug discovery and pharmaceutical research. The latest release features more than seventeen thousand drug entries, including FDA-approved small molecule drugs,  FDA-approved biotech (protein/peptide) drugs,  nutraceuticals,  and experimental drugs. For each drug entry, DrugBank contains chemical, pharmacological, and pharmaceutical properties of the drug as well as links to external databases. In addition, DrugBank also provide sequence, structure, and pathway information of around six thousand unique proteins, which are  drug targets/enzymes/transporters/carriers associated with these drugs. The information about drug structures, indications, drug-target interactions,  and pathways can support a wide range of tasks such as drug property prediction, drug activity analysis,  drug repurposing, and drug-target interaction prediction. 

\textbf{PubChem}~\cite{kim2025pubchem} is another comprehensive database of chemical molecules and their activities against biological assays, which is maintained by the National Center for Biotechnology Information (NCBI). It serves as a comprehensive resource for information on the chemical structures, properties, biological activities, and toxicity of small molecules, and is widely used in cheminformatics, bioinformatics, and computational drug discovery. PubChem is organized into three main interlinked databases: PubChem Compound, PubChem Substance, and PubChem BioAssay (PCBA). PubChem Compound database contains information of more than 100 million pure and characterized chemical compounds. The Substance section collects information of substances, including mixtures and uncharacterized substances, submitted by various data contributors. The BioAssay section contains bioactivity results from approximately 1.67 million biological assay experiments. PubChem Compound IDs are widely used across chemical databases for consistent referencing.

\textbf{MoleculeNet}~\cite{wu2018moleculenet} is a benchmarking platform designed to facilitate the development and evaluation of machine learning models for molecular property prediction. The authors curated a wide variety of datasets from other primary sources, covering different molecular properties and tasks. Although it is not as complex as the two database mentioned earlier, it had been utilized frequently in evaluating newly proposed machine learning approaches because the datasets were constructed for specific tasks and were organized in a very simple format for download. Briefly, the datasets cover four different types of properties, including Quantum Mechanics (including datasets QM7, QM8, QM9), Physical Chemistry (datasets ESOL, FreeSolv, Lipophilicity), Biophysics (datasets PCBA, MUV, HIV, BACE), and Physiology (datasets BBBP, Tox21, SIDER, ClinTox). The prediction tasks can be either classification or regression. As a reference, we provide a very brief summary for each of the datasets. 

\textbf{QM7, QM8, QM9} provide quantum mechanical properties and 3D molecular geometries that can be used as training data for quantum property prediction. QM7 includes 7,165 molecules  computed atomization energies and Coulomb matrices.  QM8 includes 21,786 molecules with calculated electronic spectra. QM9 expands to over 133,000 stable organic compounds with detailed quantum mechanical properties including  energies, geometries, and vibrational frequencies.

\textbf{ESOL} is small dataset of 1,128 molecules in SMILES format offering water solubility data, useful for evaluating solubility predictions. \textbf{FreeSolv} contains  642 small molecules with both experimental and computed hydration free energies.  \textbf{Lipophilicity} reports log D values of the octanol--water distribution coefficients for over 4,200 drug molecules, reflecting membrane permeability and solubility.

\textbf{PCBA} contains activity profiles for over 400,000 molecules against specific enzymes, receptors, and pathways, derived from PubChem BioAssay database. \textbf{MUV} is a filtered subset of PubChem BioAssay, designed to validate virtual screening techniques and includes 17 benchmark tasks. \textbf{HIV} contains more than 41 thousand molecules labeled for their ability to inhibit HIV replication based on biological assay data. \textbf{BACE} includes 1,513 inhibitors of human $\beta$-secretase 1 (BACE-1), with both binary activity labels and IC50 values.

\textbf{BBBP} includes more than two thousand molecules  labeled based on whether they can cross the blood-brain barrier. \textbf{Tox21} contains toxicity data for close to eight thousand compounds across 12 targets, used in toxicology modeling. \textbf{ToxCast} extends Tox21, with bioactivity measurements on 617 biological targets for 8,576 compounds. \textbf{SIDER} focuses on close to fifteen hundred marketed drugs and their recorded adverse drug reactions (more than five thousand side effects). \textbf{ClinTox} contains approved drugs and compounds that failed clinical trials due to toxicity concerns.

Each dataset is accompanied by task definitions (e.g., classification or regression), standard metrics (e.g., ROC-AUC, RMSE), and data preprocessing techniques (e.g., scaffold splits, random splits) to promote consistent model evaluation.

\subsection{Clinical Databases}
Clinical databases are those databases that contain clinical and health information from patients that can be used for disease prediction, treatment outcome modeling, as well as drug recommendation and precision medicine. We therefore list two popular databases here:  Medical Information Mart for Intensive Care database (MIMIC-III~\cite{johnson2016mimic} and MIMIC-IV~\cite{Johnson2023}), and the UK Biobank database~\cite{ukbiobank}.

\textbf{MIMIC-III}~\cite{johnson2016mimic} and \textbf{MIMIC-IV}~\cite{ Johnson2023} are freely accessible, large-scale clinical databases developed by the MIT Lab for Computational Physiology. While MIMIC-IV is an updated and improved version of MIMIC-III, and there are overlapped samples in the two database,  MIMIC-IV  does not encompass all the data present in MIMIC-III. We briefly discuss both databases. MIMIC-III contains de-identified health-related data associated with over 40,000 patients who stayed in critical care units of the Beth Israel Deaconess Medical Center (BIDMC) between 2001 and 2012. The database includes a wide range of data types across 26 tables, such as demographics, vital signs, laboratory test results, medications, diagnostic codes (ICD-9), procedures, imaging reports, and clinical notes. Its structured and time-stamped data makes it especially valuable for developing and validating models for disease risk and patient trajectory prediction. In addition, the dataset has also been frequently employed in studies on drug recommendations  and drug combination recommendations after extensive preprocessing. 
On the other hand MIMIC-IV includes detailed, de-identified clinical data for 180,733 patients for hospital admissions and 50,920 patients for ICU admissions from BIDMC between 2008 and 2019. Information available includes patient measurements, orders, diagnoses, procedures, treatments, and de-identified free-text clinical notes. Both datasets support a wide array of research studies and help to reduce barriers to conducting clinical research using patient level data.

\textbf{UK Biobank}~\cite{ukbiobank} is another large-scale biomedical database and research resource containing in-depth genetic, lifestyle, and health information from approximately 500,000 volunteer participants aged 40–69 at the time of recruitment (2006–2010) across the United Kingdom. It is managed by a charitable organization and made available to approved researchers for health-related research. The dataset includes a broad array of data types, such as genotyping and whole-genome sequencing, biochemical assays, physical measures, imaging data (e.g., MRI, CT scans), hospital and primary care records, and detailed lifestyle and demographic questionnaires. UK Biobank is particularly valuable for population-level studies on complex diseases. The integration of genetic with phenotypic and clinical data makes it one of the most important resources for identifying drug targets, predicting drug effects, and accelerating drug development.

\subsection{Structural Information Databases}
Structural information datasets focus on providing 3D structures and/or conformers of isolated ligand and/or protein–ligand complexes. These resources combine experimental structures with computationally refined conformations to support a range of applications -- from physics-based simulations such as free energy calculations to data-driven machine learning models that predict binding affinity or molecular properties. In this subsection, we highlight four widely used structural datasets: \textbf{ZINC}~\cite{irwin2020zinc22}, \textbf{GEOM}~\cite{axelrod2022geom}, \textbf{MISATO}~\cite{liu2024multigrained}, and \textbf{CrossDocked}~\cite{francoeur2020three}.

\textbf{ZINC}~\cite{irwin2020zinc22} is a meticulously curated repository containing over 230 million commercially accessible compounds. It includes 3D structures, physicochemical properties, and vendor metadata. Unlike theoretical libraries, ZINC focuses on experimentally testable molecules, facilitating streamlined drug discovery workflows. The database provides SMILES strings, 3D structures, and drug-like property classifications, organized into tranches for targeted virtual screening. As a benchmark for docking and virtual screening studies, ZINC accelerates the transition from computational predictions to experimental validation, serving as a critical tool for hit identification and lead optimization.

\textbf{GEOM}~\cite{axelrod2022geom} contains quantum mechanics-optimized 3D geometries for approximately 30 million conformers representing 450,000 drug-like molecules. Each structure is refined using density functional theory (DFT) to capture realistic conformational landscapes, emphasizing ensembles of energetically feasible states rather than static geometries. This ensemble approach advances protein-ligand interaction modeling, conformer generation algorithms, and force field validation. Its high accuracy has substantially advanced 3D-aware machine learning models for molecular property prediction and molecule generation.

\textbf{MISATO}~\cite{liu2024multigrained} is a machine learning-oriented dataset comprising roughly 20,000 experimentally resolved protein–ligand complexes. Each complex undergoes structural refinement and quantum-mechanical optimization to address stereochemical, geometric, and protonation inconsistencies. Around 17,000 complexes are further subjected to explicit-solvent molecular dynamics (MD) simulations. This dynamic data captures conformational flexibility and binding pocket dynamics. In addition, MISATO includes quantum-derived electronic descriptors, partial charges, and preprocessing utilities tailored for machine learning pipelines. By integrating static structures with time-resolved dynamics, MISATO enables modeling of transient binding states and induced-fit effects, overcoming the limitations of single-conformation datasets.

\textbf{CrossDocked}~\cite{francoeur2020three} curates 18,450 non-redundant protein–ligand complexes derived from the Protein Data Bank (PDB)~\cite{berman2000protein}, using a systematic cross-docking approach. Ligands are docked into non-cognate, structurally similar binding pockets to produce a diverse set of over 22.5 million binding poses. The dataset features cluster-based predefined splits for evaluating model generalizability to unseen targets and provides dual metrics for assessing both pose accuracy and binding affinity predictions. By modulating the structural similarity between docking receptors and their native counterparts, CrossDocked enables rigorous evaluation of docking algorithms under realistic scenarios where exact receptor structures may be unknown. Designed as a comprehensive benchmark, CrossDocked supports the standardized training and evaluation of 3D CNNs and other ML models for non-native protein–ligand interaction modeling, with broad implications for virtual screening.

\subsection{Molecular Interaction Databases}
Molecular interactions databases record relationships among different molecules in various formats and are fundamental in the study of molecular biology, computational chemistry, and drug discovery. Researchers use these datasets to elucidate biochemical pathways, predict binding affinities, evaluate selectivity, and simulate off-target effects. We include datasets that capture a range of molecular interactions, grouped into two categories: protein–ligand binding and drug–drug interactions.

\subsubsection{Protein-Ligand Binding Databases}
Protein–ligand binding describes the specific interactions between proteins (often therapeutic targets or receptors) and small-molecule ligands (including drugs). These interactions drive most biochemical modulation and are critical for drug discovery, off-target prediction, and mechanistic studies. Below, we summarize several widely used public datasets.

\textbf{ChEMBL}~\cite{zdrazil2024chembl} is a manually curated database focusing on bioactive molecules with drug-like properties. To assess the binding affinity of small-molecule ligands to their targets, ChEMBL primarily uses experimental bioactivity data extracted from scientific literature. To facilitate comparison and analysis, data from different sources undergoes a standardization process so that measurement type, value, and units are comparable. In addition to binding affinity measurements, ChEMBL also contains rich information about compounds, targets, experimental assays, and original sources.  The current release of ChEMBL (release 35)  includes approximately 2.5 million distinct compounds and 21.1 million bioactivity measurements derived from 1.7 million biological assays across 16,000 biomolecular targets. 
ChEMBL supports a wide range of applications, including structure–activity relationship analysis, off-target prediction, and drug repurposing. Its datasets are available via web interfaces, APIs, and bulk downloads under open Creative Commons licenses.

\textbf{PDBbind Dataset}~\cite{liu2017pdbbind}  was created to collect experimentally measured binding data from literature for the biomolecular complexes with high-resolution 3D structures in the Protein Data Bank (PDB).  It provides an essential linkage between the energetic and structural information of those complexes, which is helpful for various computational and statistical studies on  docking validation, scoring-function development, affinity prediction, molecular recognition, and drug discovery. The most recent version (2024) was released on a commercial platform called PDBbind+ with a free demo version. It currently contains experimental binding affinity data for 27,385 protein-ligand complex, 4,594 protein-protein complex, 1.440 protein-nucleic acid complex and 234 nucleic acid-ligand complex.

\textbf{Metz Dataset}~\cite{metz2011navigating} focused exclusively on kinase inhibition activities. It contains over 150,000 kinase inhibitory measurements, comprising more than 3,800 compounds tested against 172 different protein kinases. Based on these measurements, the authors constructed a comprehensive kinome interaction network, enabling systematic analysis of kinase–inhibitor interactions. This dataset is applicable not only to binding affinity prediction but also to the design of multi-kinase inhibitors.

\textbf{Davis Dataset}~\cite{davis2011comprehensive} was developed by Davis et al. to provide a broad target panel covering over 80\% of the human catalytic kinome. It includes selectivity profiles of 72 kinase inhibitors tested against 442 human kinases. With more than 30,000 high-precision measurements obtained from a standardized binding assay, the dataset supports various tasks such as binding affinity prediction, selectivity profiling, off-target prediction, and regression model validation. The uniform experimental design and extensive kinase coverage make the dataset a preferred benchmark for visually screening compound libraries and identifying novel inhibitors.

Via a systematic evaluation of target selectivity profiles across three different biochemical assays of kinase inhibitors, Tang et al.~\cite{tang2014making} introduced a model-based approach and a unified affinity metric known as the ``KIBA score'', to integrate complementary information captured by different  bioactivity types. The resulting \textbf{KIBA Dataset} comprises a drug–target bioactivity matrix involving 52,498 chemical compounds and 467 kinase targets, with a total of 246,088 KIBA scores. This statistically harmonized dataset, designed to minimize experimental variability, has become a widely used benchmark for training machine learning models in drug–target affinity prediction.

\subsubsection{Drug–Drug Interaction Databases}
While comprehensive resources such as DrugBank include DDI information, they often require additional processing to extract pure interaction data. Furthermore, they may not include all the measurements from high-throughput screening assays, for example, dose responses for different dose combinations of drug pairs.  Below, we highlight some widely used datasets for both adverse and synergistic DDI prediction tasks. For adverse or general DDI prediction, commonly used datasets include TWOSIDES~\cite{tatonetti2012data}, Deng’s Dataset~\cite{deng2020multimodal}, and ChChMiner~\cite{biosnapnets}. 

\textbf{TwoSIDES}~\cite{tatonetti2012data} is a database of polypharmacy side effects for pairs of drugs. It was constructed by mining the U.S. Food and Drug Administration (FDA) adverse event reporting systems (FAERS)~\cite{FAERS}. The dataset consists of 868,221 statistically significant association between 59,220 drug pairs and 1,301 adverse events. Only associations that cannot be clearly attributed to either drug alone were included. It improves the detection and prediction of adverse effects of drug interactions.

\textbf{Deng’s Dataset}~\cite{deng2020multimodal} consists of 74,528 distinct drug–drug interactions among 572 approved drugs extracted from DrugBank entries by appling NLP algorithms. Each interaction is recorded as a four-element tuple: (\emph{drug A}, \emph{drug B}, \emph{mechanism}, \emph{action}), where the `mechanism' means the effect of drugs in terms of the metabolism, the serum concentration, the therapeutic efficacy and so on. The `action' represents increase or decrease. The categorization of interactions into different types of mechanism of actions  is useful for understanding the actual logic hidden behind the combined drug usage or adverse reactions and for evaluating algorithms that aim to recover them. 

\textbf{ChChMiner} is a sub-dataset from the Stanford Biomedical Network Dataset Collection (BioSNAP)~\cite{biosnapnets}, which itself is a comprehensive database providing information about relationships between a variety of biological, chemical, or clinical entities including drugs. ChChMiner is a network of interactions among 1,514 FDA-approved drugs, extracted from drug labels, scientific publications, and DrugBank, with 48,514 drug–drug interactions.

For synergistic DDI prediction, we include an aggregated dataset from various sources (DrugCombDB~\cite{Liu2019}) and two high-throughput screening data of drug pairs on cancer cell lines (O'Neil’s Dataset~\cite{ONeil2016} and AstraZeneca’s Dataset~\cite{menden_community_2019}). \textbf{DrugCombDB}~\cite{Liu2019} aggregated drug combination data from various data sources: high-throughput screening assays of drug combinations, manual curations from the literature, FDA-approved therapies and failed clinical trials, as well as some earlier drug combination databases such as DCDB~\cite{liu2014dcdb}. The database comprises 448,555 drug combinations involving 2,887 unique drugs and 124 human cancer cell lines. In addition, DrugCombDB has more than 6,000,000 quantitative dose responses from which multiple synergy scores to determine the overall synergistic or antagonistic effects of drug combinations were calculated based on different models.  As a comprehensive database with a large number of drug combinations, DrugCombDB would greatly facilitate and promote the discovery of novel synergistic drugs for the therapy of complex diseases. 

By using a high-throughput platform for unbiased identification of synergistic and additive drug combinations, O’Neil et al. ~\cite{ONeil2016} created a dataset comprising 38 experimental or approved drugs tested across 39 diverse cancer cell lines, using a 4-by-4 dosing regimen for a total of 583 drug–drug combinations. More recently, a similar but larger dataset was created by AstraZeneca and shared with the research community through a DREAM Challenge~\cite{menden_community_2019}, which consisted of 11,576 experiments from 910 combinations of 118 drugs across 85 molecularly characterized cancer cell lines. Both datasets can serve as benchmark datasets and can accelerate the development of computational approaches for drug combination synergy prediction.

\subsection{Challenges in Data}

Deep learning models require large volumes of high-quality data to effectively capture the complex relationships between molecular entities and to predict properties of chemical compounds. However, several limitations still hinder progress. First of all, substantial heterogeneity exists in different databases. Data often comes in different formats with different identifiers, making integration and standardization difficult. Data derived from different labs or measurement techniques may not be comparable, limiting transferability. In the extreme cases, different sources may report contradictory effects or properties for the same compound. 

Secondly, considerable biases exist in different data sources. Almost all the databases have the coverage bias: most datasets focus on well-studied proteins, pathways, diseases, and only a small portion of theoretically limitless set of all possible chemical compounds, leaving gaps in knowledge for novel targets, rare diseases, and novel drug candidates. Underrepresented molecular classes such as biotech medicines/biologics are often excluded due to lack of structured data. In addition, many datasets consist of label biases, including imbalance labels, and noisy and incomplete labels. Many datasets consist of only positive (or negative) samples because of the nature of the datasets. For example, drug-drug interactions were mostly reported when side effects were observed, which only provided positive samples. On the contrary, failed clinical trials only contained negative results. Label imbalance makes the prediction challenging and motivates the development of pre-training and few-shot learning algorithms. Some benchmark datasets, especially those with smaller sizes, have dataset specific biases. Models trained on those datasets may suffer from overfitting and not generalize well.

Finally, there are limitations on the scopes of data. Many benchmark datasets are static and not routinely updated with new discoveries. Furthermore, due to privacy concern and regulatory and ethical constraints, individual level, patient-centric data is not accessible to the broad research community. In particular, few datasets combine molecular data with real-world patient electronic health record (EHR) data, hindering the progress in personalized medicine.

\section{Discussion and Conclusion}
In recent years, deep learning approaches -- particularly GNNs -- have garnered increasing attention in the field of computational drug discovery. In this survey, we systematically review studies published since 2021, highlighting the significant role GNNs have played across three core application areas: molecule generation, molecular property prediction, and drug–drug interaction prediction. We examine how GNNs effectively model chemical structures and capture complex molecular patterns, discussing the strengths and limitations of current approaches, along with the major challenges faced by the research community. The papers included in this review represent state-of-the-art advancements in the field and demonstrate that molecular graph representation learning has become a dominant paradigm across these applications.

Specifically, we observe that GNNs have been instrumental in advancing molecule generation by enabling the design of novel compounds with desired properties through both unconstrained and constrained generation strategies. In the area of molecular property prediction, the shift towards utilizing 3D molecular graphs has led to more accurate and robust outcomes, particularly when combined with message-passing mechanisms and contrastive learning techniques. For DDI prediction, GNNs have opened promising avenues in personalized medicine by identifying safe and effective drug combinations tailored to individual patient profiles. This is especially impactful for drug repurposing, where GNNs can significantly accelerate the development of combinatorial therapies.

Across the reviewed studies, several common trends emerge. First, pre-training and self-supervised learning have become widespread, substantially enhancing the performance of GNN-based models. These techniques are particularly effective in mitigating the issue of limited labeled data and contribute to the development of more generalizable models.
Second, the incorporation of domain-specific knowledge into GNN architectures has led to noticeable improvements in model performance, suggesting a movement toward more specialized and biologically informed models. Third, many recent works adopt multi-modal approaches that integrate diverse input formats, including 2D and 3D molecular graphs, as well as SMILES strings, sometimes combining GNNs with other deep learning architectures. This fusion of complementary molecular representations enables a more comprehensive understanding of the data and has the potential to significantly enhance predictive performance.

Despite these advancements, several challenges remain, which also point to promising future directions. First, as previously discussed, data scarcity and the limited availability of high-quality, diverse datasets continue to constrain the full potential of GNNs in drug discovery. Second, the interpretability of GNN models remains a critical hurdle that must be addressed to foster trust and facilitate their adoption in real-world applications. As GNN architectures grow increasingly complex, ensuring interpretability is paramount. As emphasized by Henderson et al.~\cite{henderson2021improving}, future research should prioritize models that not only provide accurate predictions but also offer clear explanations for their decisions. Bridging this gap between computational predictions and human understanding will be essential for generating actionable scientific insights. Lastly, multi-omics integration approaches combine data from various biological levels (genomics, transcriptomics, proteomics, metabolomics,  epigenomics, etc.) to create a more comprehensive understanding of disease mechanisms. Integration of GNN-based models with multi-omics data holds significant promise and could further revolutionize the landscape of drug discovery by identifying better drug targets, developing personalized therapies, and eventually improving treatment outcomes.

\section{Competing Interests}
No competing interest is declared.

\section{Author Contributions Statement}
Zhengyu Fang: Conceptualization, Writing – Original Draft.
Xiaoge Zhang: Writing – Original Draft, Writing – Review \& Editing.
Anyin Zhao: Writing – Original Draft, Writing – Review \& Editing.
Xiao Li: Supervision.
Huiyuan Chen: Conceptualization, Investigation.
Jing Li: Supervision, Conceptualization, Writing – Review \& Editing.
All authors reviewed and approved the final manuscript.

\section{Acknowledgments}
This work is supported in part by NSF CCF-2200255, NSF CCF-2006780, NSF IIS-2027667, NIH U01AG073323, NIH R01HG009658, NIH 1R01HL159170 and NIH 1R01NR02010501.

\bibliographystyle{naturemag}
\bibliography{reference}


\newpage
\section{Appendix}
In Table~\ref{tab:approach_url} we provide the URLs of the approaches we summarized in Table~\ref{tab:methods}, however, some of approaches do not offer URLs.

\begin{table*}[h!]
\small
\centering
\caption{There are the URLs for the codes of methods.}
\label{tab:approach_url}
\resizebox{0.75\textwidth}{!}{
\begin{tabular}{l|l}
\toprule
\textbf{Methods} & \textbf{URLs} \\
\midrule
ConfVAE~\cite{xu2021endtoend} & \url{https://github.com/MinkaiXu/ConfVAE-ICML21} \\
CGIB~\cite{lee2023conditional} & \url{https://github.com/Namkyeong/CGIB} \\
VonMisesNet~\cite{swanson2023von} & \url{https://github.com/thejonaslab/vonmises-icml-2023} \\
MGraphDTA~\cite{yang2022mgraphdta} & \url{https://github.com/guaguabujianle/MGraphDTA} \\
MoLeR~\cite{maziarz2022learning} & \url{https://github.com/microsoft/molecule-generation} \\
SphereNet~\cite{liu2022spherical} & \url{https://github.com/divelab/DIG} \\
MolKGNN~\cite{liu2023interpretable} & \url{https://github.com/meilerlab/MolKGNN} \\
MiCam~\cite{geng2023novo} & \url{https://github.com/MIRALab-USTC/AI4Sci-MiCaM} \\
GraphMVP~\cite{liu2021pre} & \url{https://github.com/chao1224/GraphMVP} \\
AR~\cite{luo20213d} & \url{https://github.com/luost26/3D-Generative-SBDD} \\
3D-Informax~\cite{stark20223d} & \url{https://github.com/HannesStark/3DInfomax} \\
GraphBP~\cite{liu2022generating} & \url{https://github.com/divelab/GraphBP} \\
UnifiedPML~\cite{zhu2022unified} & \url{https://github.com/teslacool/UnifiedMolPretrain} \\
Pocket2Mol~\cite{peng2022pocket2mol} & \url{https://github.com/pengxingang/Pocket2Mol} \\
FLAG~\cite{zhang2023molecule} & \url{https://github.com/zaixizhang/FLAG} \\
MoleculeSDE~\cite{liu2023group} & \url{https://github.com/chao1224/MoleculeSDE} \\
SQUID~\cite{adams2022equivariant} & \url{https://github.com/keiradams/SQUID} \\
3D-PGT~\cite{wang2023automated} & \url{https://github.com/LARS-research/3D-PGT} \\
SG-CNN~\cite{jones2021improved} & \url{https://github.com/llnl/fast} \\
IGN~\cite{jiang2021interactiongraphnet} & \url{https://github.com/zjujdj/IGN} \\
O-GNN~\cite{zhu2022mathcal} & \url{https://github.com/O-GNN/O-GNN} \\
MP-GNN~\cite{li2022multiphysical} & \url{https://github.com/Alibaba-DAMO-DrugAI/MGNN} \\
MoleOOD~\cite{yang2022learning} & \url{https://github.com/yangnianzu0515/MoleOOD} \\
GraphscoreDTA~\cite{wang2023graphscoredta} & \url{https://github.com/CSUBioGroup/GraphscoreDTA} \\
MGSSL~\cite{zhang2021motif} & \url{https://github.com/zaixizhang/MGSSL} \\
NERE~\cite{jin2023unsupervised} & \url{https://github.com/wengong-jin/DSMBind} \\
MoCL~\cite{sun2021mocl} & \url{https://github.com/illidanlab/MoCL-DK} \\
KCL~\cite{fang2022molecular} & \url{https://github.com/ZJU-Fangyin/KCL} \\
FABind~\cite{pei2023fabind} & \url{https://github.com/QizhiPei/FABind} \\
MCHNN~\cite{liu2023multi} & \url{https://github.com/Liuluotao/MCHNN} \\
HiMol~\cite{zang2023hierarchical} & \url{https://github.com/ZangXuan/HiMol} \\
MHNfs~\cite{schimunek2023context} & \url{https://github.com/microsoft/FS-Mol} \\
GS-Meta~\cite{zhuang2023graph} & \url{https://github.com/HICAI-ZJU/GS-Meta} \\
MRCGNN~\cite{Xiong2023} & \url{https://github.com/Zhankun-Xiong/MRCGNN} \\
DPDDI~\cite{feng2020dpddi} & \url{https://github.com/NWPU-903PR/DPDDI} \\
SRR-DDI~\cite{NIU2024111337} & \url{https://github.com/NiuDongjiang/SRR-DDI} \\
DAS-DDI~\cite{NIU2024104672} & \url{https://github.com/NiuDongjiang/DAS-DDI} \\
SSF-DDI~\cite{Zhu2024} & \url{https://github.com/ZHJING25/SSF-DDI} \\
DeepDDS~\cite{Zhu2024} & \url{https://github.com/Sinwang404/DeepDDS/tree/master} \\
SafeDrug~\cite{yang2021safedrug} & \url{https://github.com/ycq091044/SafeDrug} \\
Geo-DEG~\cite{guo2023hierarchical} & \url{https://github.com/gmh14/Geo-DEG} \\
MoleRec~\cite{yang2023molerec} & \url{https://github.com/yangnianzu0515/MoleRec} \\
DVMP~\cite{zhu2023dual} & \url{https://github.com/microsoft/DVMP} \\
Carmen~\cite{Chen2023} & \url{https://github.com/bit1029public/Carmen} \\
\bottomrule
\end{tabular}}
\end{table*}

\end{document}